\documentclass{llncs}
\usepackage{graphicx}
\usepackage{amssymb}
\usepackage{amsmath}
\usepackage{booktabs}
\usepackage{hyperref}
\usepackage{bm}
\usepackage{bbm}

\usepackage{todonotes}

\newcommand{\beq}{\begin{equation}}
\newcommand{\eeq}{\end{equation}}

\DeclareMathOperator*{\argmin}{arg\,min}

\begin{document}


\title{Registration by Regression (RbR): a framework for interpretable and flexible atlas registration}


\author{Karthik Gopinath\thanks{Corresponding author: K. Gopinath. \textbf{Email:}~\email{kgopinath@mgh.harvard.edu}}\inst{1} \textsuperscript{*} \and
Xiaoling Hu\inst{1}\textsuperscript{*} \and 
Malte Hoffmann\inst{1} \and
Oula Puonti\inst{1,2}\textsuperscript{**} \and
Juan Eugenio Iglesias\inst{1,3,4}\textsuperscript{,**} }

\institute{Massachusetts General Hospital and Harvard Medical School \and
Danish Research Centre for Magnetic Resonance, Copenhagen University Hospital \and Centre for Medical Image Computing, University College London \and Computer Science and AI Laboratory, Massachusetts Institute of Technology\\
\textsuperscript{*}Contributed equally as first authors and \textsuperscript{**}contributed equally as senior authors.}


\maketitle

\vspace{-1mm}
\begin{abstract}

In human neuroimaging studies, atlas registration enables mapping MRI scans to a common coordinate frame, which is necessary to aggregate data from multiple subjects. 
Machine learning registration methods have achieved excellent speed and accuracy but lack interpretability and flexibility at test time (since their deformation model is fixed). More recently, keypoint-based methods have been proposed to tackle these issues, but their accuracy is still subpar, particularly when fitting nonlinear transforms. Here we propose Registration by Regression (RbR), a novel atlas registration framework that: is highly robust and flexible; can be trained with cheaply obtained data; and operates on a single channel, such that it can also be used as pretraining for other tasks.  RbR predicts the $(x,y,z)$ atlas coordinates for every voxel of the input scan (i.e., every voxel is a keypoint), and then uses closed-form expressions to quickly fit transforms using a wide array of possible deformation models, including affine and nonlinear  (e.g., Bspline, Demons, invertible diffeomorphic models, etc.). Robustness is provided by the large number of voxels informing the registration and can be further increased by robust estimators like RANSAC. Experiments on independent public datasets show that RbR yields more accurate registration than competing keypoint approaches, over a wide range of deformation models. 


\end{abstract}

\section{Introduction}

Image registration seeks to find a spatial mapping between two images, and is a crucial component of human neuroimaging pipelines, e.g., for measuring change between different timepoints of the same subject~\cite{holland2011nonlinear}; building population atlases or subject-specific templates via co-registration of multiple scans~\cite{joshi2004unbiased}; measuring structural differences via tensor-based morphometry~\cite{hua2009optimizing}; mapping pre-, intra-, and post-operative images~\cite{alam2018medical}; or automated segmentation~\cite{iglesias2015multi}.
One important application of these methods is registering brain MRI scans to digital atlases, such as MNI. Atlas registration enables aggregation of spatial information of multiple subjects into a common coordinate frame (e.g., for voxel-based morphometry~\cite{ashburner2000voxel}), as well as mapping of population-wide information to the native space of a specific subject, e.g., label priors for image segmentation~\cite{fischl2002whole}. Atlas registration typically relies on the same methods as pairwise registration.

Most classical pairwise registration methods use numerical optimization~\cite{nocedal1999numerical} to minimize a cost function, often combining an image similarity term with a regularizer; the former seeks to align the images to each other, while the latter prevents excessively convoluted (unrealistic) deformation. Decades of research have provided medical image registration with features such as diffeomorphisms~\cite{avants2008symmetric}, symmetry~\cite{christensen2001consistent}, or inter-modality mapping~\cite{pluim2003mutual}. 

In 2017, faster registration methods based on deep learning were initially published. The first attempts were supervised and sought to predict ground truth fields obtained, e.g., with accurate classical methods~\cite{yang2017quicksilver}. These algorithms then evolved into unsupervised methods, trained with the same losses used by classical algorithms  -- possibly augmented with segmentation losses~\cite{balakrishnan2019voxelmorph,de2019deep}. By now, deep learning registration methods have incorporated many of the features of their classical counterparts, e.g., diffeomorphisms~\cite{krebs2019learning}, symmetry~\cite{iglesias2023ready}, progressive warping~\cite{lv2022joint}, or inter-modality support~\cite{hoffmann2021synthmorph}.
However, deep learning registration also has drawbacks, including:
lack of interpretability (as many other deep learning domains);  lack of flexibility of the deformation model, which is fixed and cannot be modified at test time (beyond tweaking the regularizer weight with hypernetworks~\cite{hoopes2022learning}); and  lack of robustness of the affine component.

Very recently, keypoint methods~\cite{wang2023robust,hoffmann2023affine} have been proposed, which predict blob-like feature maps such that their centers of gravity can be used to analytically compute a transform between the two images. While the feature prediction remains opaque, this approach partly address the lack of interpretability by providing insight into the factors driving the alignment, i.e., the keypoints. 
Keypoints also have potential in dynamic selection of deformation model at test time to warp one point cloud to the other, e.g., affine or thin-plate spline (TPS) transforms in KeyMorph~\cite{wang2023robust}. 
Finally, keypoints also effectively tackle the lack of robustness of the affine component. Earlier learning approaches for nonlinear registration assumed that the inputs were already affinely aligned; some tried to predict the 12-parameter affine matrix~\cite{de2019deep,mok2022affine}, but these predictions were generally sensitive to initialization and often failed to generalize well to new datasets. 

Here we present Registration by Regression (RbR), a novel framework for flexible, robust registration of brain MRI scans to atlases. RbR is an evolution of keypoint methods where \emph{every voxel} is a keypoint: it is a supervised convolutional neural network (CNN) that predicts $(x,y,z)$ atlas coordinates for every input voxel. RbR has a number of advantages: 
\textit{(i)}~Since the number of keypoints is very high ($\sim 10^6$ vs $\sim 10^2$ in KeyMorph), nonlinear transforms can be fitted more accurately. 
\textit{(ii)}~After a forward pass of the CNN,  a wide array of popular transforms and regularizers can be fit in closed-form,  including affine and nonlinear models like Bsplines~\cite{rueckert1999nonrigid}, Demons~\cite{thirion1998image}, or the diffeomorphic log-polyaffine model~\cite{arsigny2009fast}, among others. 
\textit{(iii)}~It supports robust estimators like RANSAC~\cite{fischler1981random}. 
\textit{(iv)}~Every hyperparameter of the deformation model can be readily modified; this is in contrast with hypernetwork approaches~\cite{hoopes2022learning}, where only specific parameters can be tuned as the space of deformations is constant (e.g., control point spacing cannot be specified). 
\textit{(v)}~Being a prediction on a single image (channel), it can be directly and easily used as pretraining (``feature extractor'') for other tasks.
\textit{(vi)}~Compared with KeyMorph, it does not need pretraining to avoid clustering of keypoints in the center of the image. 
\textit{(vii)}~Compared with classical registration, it is much faster (particularly when  considering that one fit different transforms ``for free'' after the forward pass) and can be used as building block in deep learning architectures (as a frozen differentiable block to align images, or as a feature extractor as mentioned above).


\section{Methods}

\subsection{Data preparation}

\begin{figure}
    \centering
    \includegraphics[width=\textwidth]{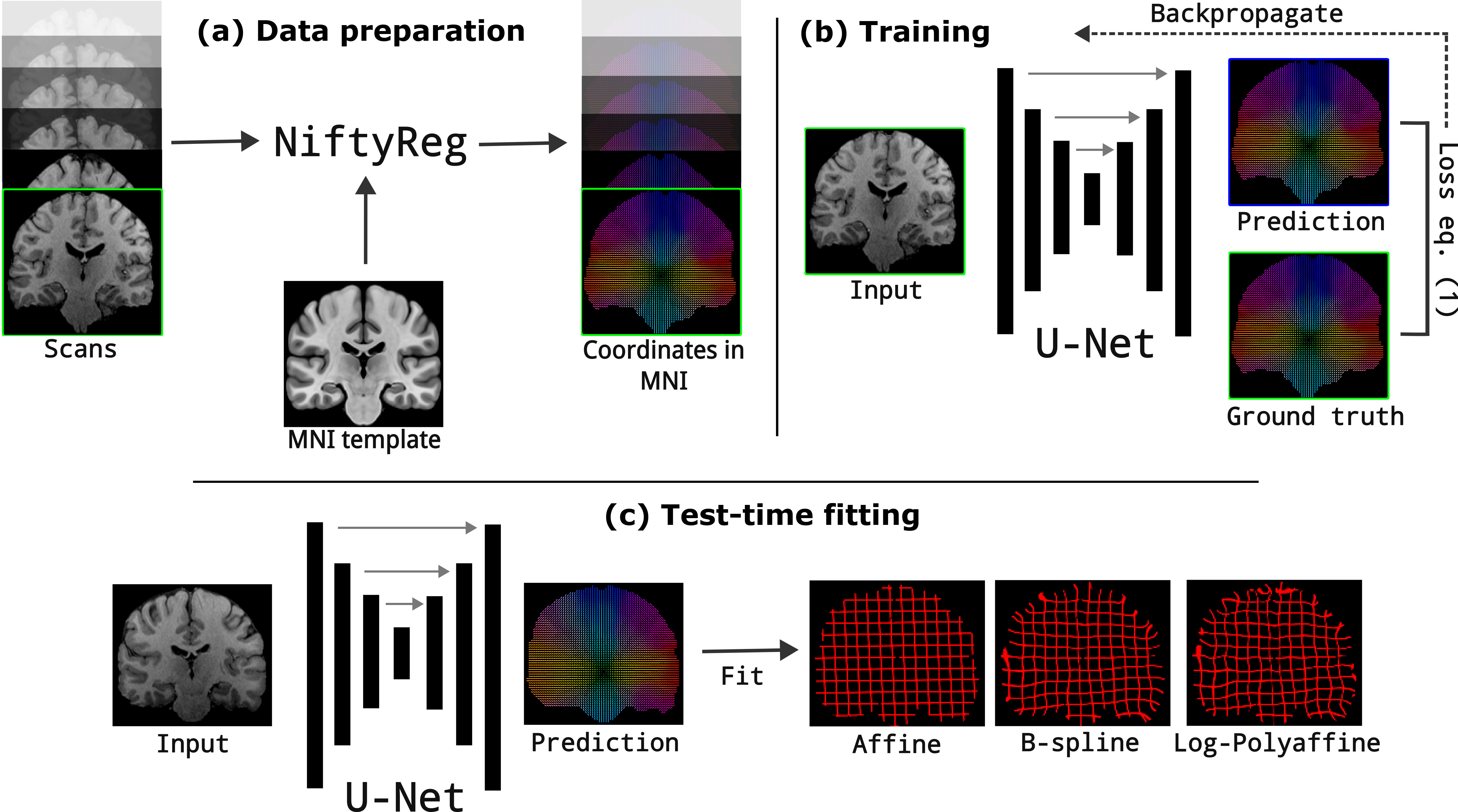}
    \caption{(a)~Training data are prepared by nonlinearly registering T1w scans from HCP and ADNI to the MNI template using NiftyReg. (b)~A U-Net is trained to predict the MNI coordinates for every voxel of an input scan. (c)~At test time, a wide array of transformations can be fitted between the input scan and predicted MNI coordinates.}
    \label{fig:overview}
\end{figure}

RbR (Figure~\ref{fig:overview}) is a regression CNN that estimates, for every voxel of the input scan (with discrete coordinates $x,y,z$), its corresponding atlas coordinates $x',y',z'$. RbR uses supervised training with accurate nonlinear deformations obtained with a classical registration approach $\mathcal{R}$. If $I(x,y,z)$ is the input image and $A$ is an atlas, the ground truth registration is:
$$
\mathbf{F}: \mathbb{R}^3 \rightarrow \mathbb{R}^3, \text{where} \quad \mathbf{F}(x, y, z) = [x', y', z'] = \mathcal{R}(x, y, z; I, A).
$$
Therefore, a large labeled dataset can be obtained ``for free'' by simply registering a set of unlabeled images to the atlas at hand using a classical registration algorithm -- ideally a diffeomorphic methods that yields smooth and invertible fields ~\cite{avants2008symmetric}.
We note that unlabeled images exist in abundance in public datasets -- particularly 1mm isotropic MPRAGEs typically used in neuroimaging research.

We also note that, compared with using synthetic fields, the performance of our strategy is bounded by that of the classical algorithm $\mathcal{R}$. However, synthesis of realistic \textit{(atlas, image, deformation)} triplets is not trivial, since atlases are much smoother than images. Moreover, classical algorithms still provide state-of-the-art accuracy in atlas registration -- speed limitations aside.

\subsection{Training}

Given a 3D voxel-wise regression CNN $\bm{N}(\cdot)$ with parameters $\bm{\theta}$, an atlas $A$, and a set of $J$ images $\{I_j\}_{j=1,\ldots,J}$,  the goal is to minimize the loss:
\begin{equation}
\hat{\bm{\theta}} = \argmin_{\bm{\theta}} \;  \sum_{j=1}^J \frac{1}{|\Omega_j|}\sum_{(x,y,z)\in\Omega_j} \ell_1[\mathcal{R}(x, y, z; I_k, A) - \bm{N}(x, y, z; I_k, \bm{\theta})],
\label{eq:loss}
\end{equation}
where $\Omega_j$ is the foreground of $I_j$ (i.e., voxels inside the brain),  
and we use $\ell_1$ rather than $\ell_2$ due to its higher robustness. We note that no spatial regularization is needed: the CNN just does its best in a voxel-by-voxel basis, and smoothness is provided by the deformation model and its parameters at test time. Nevertheless, augmentation is crucial to endow the trained models with the best possible generalization ability. We use aggressive augmentation during training, including random blurring, bias field, noise, and deformation (affine and elastic).

\subsection{Test-time fitting}

Given a CNN prediction $[x',y',z'] = \bm{N}(x, y, z; I, \hat{\bm{\theta}})$ (which does \emph{not} require the atlas), one can fit a deformation model to $[x',y',z']$ to obtain the final atlas registration. We consider three different types of models in this article: a family of models based on basis functions, a model based on the Demons algorithm~\cite{thirion1998image}, and a diffeomorphic model based on the log-polyaffine framework~\cite{arsigny2009fast}. 

\smallskip\noindent\textbf{Basis functions: } a large class of transforms  can be written as a sum of spatial basis functions and can be fitted to $[x',y',z']$ using regularized least squares, one spatial coordinate at the time. Let $\bm{d}$ be the flattened column vector of $x'$, $y'$, or $z'$ coordinates. Let $\bm{\phi}$ be a set of $B$ spatial functions evaluated on the discrete image domain $\Omega$, such that each row corresponds to a spatial location (flattened), and each column corresponds to a basis function. Finally, let $\bm{c}$ be a column vector with $B$ elements, corresponding to the coefficients of the basis functions. Then, the fitted transform is  $\bm{\phi}\bm{c}$ and its squared coordinate error is:
$$
E_{coord} = [\bm{d} - \bm{\phi} \bm{c}]^t [\bm{d} - \bm{\phi} \bm{c}] = \bm{d}^t \bm{d} + \bm{c}^t \bm{\phi}^t \bm{\phi} \bm{c} - 2 \bm{d}^t \bm{\phi} \bm{c}.
$$
This fit can be regularized with quadratic penalty terms while remaining closed form. Here we choose the membrane energy, which penalizes the squared norm of the gradient of the deformation field. Let $\bm{G}_x, \bm{G}_y, \bm{G}_z$ be the spatial gradients of the basis functions; as for $\bm{\phi}$, every row corresponds to a spatial location, and every column to a basis function. The regularizer is then given by:
$$
E_{reg} = [\bm{G}_x \bm{c}]^t [\bm{G}_x \bm{c}]  + [\bm{G}_y \bm{c}]^t [\bm{G}_y \bm{c}]  + [\bm{G}_z \bm{c}]^t [\bm{G}_z \bm{c}] = \bm{c}^t [\bm{G}^t_x \bm{G}_x + \bm{G}^t_y \bm{G}_y + \bm{G}^t_z \bm{G}_z]  \bm{c}.
$$
$E_{coord}$ and $E_{reg}$ are combined into an objective function $E$ using a relative weight $\lambda$. Setting the gradient to zero yields the regularized least square estimate:
\begin{align}
\nabla E &= \nabla (E_{coord}   +\lambda E_{reg})  = 2 \bm{\phi}^t \bm{\phi} \bm{c} - 2 \bm{\phi}^t \bm{d} + 2 \lambda [\bm{G}^t_x \bm{G}_x + \bm{G}^t_y \bm{G}_y + \bm{G}^t_z \bm{G}_z] \bm{c} = 0,\nonumber\\
& \Rightarrow \bm{c} = [\bm{\phi}^t \bm{\phi} + \lambda (\bm{G}^t_x \bm{G}_x + \bm{G}^t_y \bm{G}_y + \bm{G}^t_z \bm{G}_z)]^{-1} \bm{\phi}^t \bm{d}\nonumber.
\end{align}
Within this family of transforms, we consider three in this article:
\begin{itemize}
\item \textbf{Affine}: a standard affine transform, fitted with all available voxels. We note that, in this case, the basis function matrix is simply $\bm{\phi} = [\bm{x}, \bm{y}, \bm{z}, \mathbbm{1}]$ (where $\mathbbm{1}$ is the all-ones vector), and the gradients of the basis functions are ignored. 
\item \textbf{Affine-RANSAC}: an affine transform fitted robustly with RANSAC. 
\item \textbf{Bsplines}~\cite{rueckert1999nonrigid}: with control point spacing specified by the user.
\end{itemize}
Other basis previously used in registration that would be straightforward to implement include lower-order polynomials~\cite{woods1998automated} or the discrete cosine transform~\cite{ashburner1999nonlinear}. 

\smallskip\noindent\textbf{Demons-like model:}
The demons algorithm~\cite{thirion1998image} computes a nonlinear registration by alternating between: \textit{(i)}~estimating force vectors via an optical-flow-like algorithm~\cite{horn1981determining}; \textit{(ii)}~smoothing these force vectors with a Gaussian kernel; and \textit{(iii)}~applying the deformation before recomputing the force vectors. Here, we propose a similar algorithm with the difference that, since the output of the CNN is constant, we only take one step, i.e., we just filter the output of the CNN with a Gaussian kernel. The standard deviation is specified by the user.

\smallskip\noindent\textbf{Log-polyaffine model:}
Diffeomorphic models that can be analytically inverted are desirable in registration. Here we explore a log-polyaffine model~\cite{arsigny2009fast} that is fitted to the output of the CNN as follows. First, we subdivide the image domain into cubic supervoxels of fixed, user-defined width $W$; this parameter controls the flexibility of the transform. Next, we compute an affine transform for every supervoxel $s$, assuming that it contains a minimum number of voxels in the brain mask. If we denote this affine transform by $\bm{T}_s$ in homogeneous coordinates, 
if can be shown that~\cite{arsigny2009fast}:
$$
\log(\bm{T}_s) = \begin{pmatrix}
\bm{L}_s & \bm{v}_s\\
0 & 0
\end{pmatrix},	
$$
and it can be shown that the affine transform can be represented by a stationary velocity field (SVF) given by: $\bm{\Psi}_s(x,y,z) = \bm{v}_s + \bm{L}_s [x, y, z]^t$. The log-polyaffine framework computes a global SVF $\bm{\Psi}$ as a weighted sum of SVFs:
$$
\bm{\Psi}(x,y,z) = \sum_s w_s(x,y,z) (\bm{v}_s + \bm{L}_s [x, y, z]^t),
$$
where $w_s(x,y,z)$ are normalized non-negative weights obtained with a Gaussian function of the distance between $(x,y,z)$ and the center of supervoxel $s$; we set the standard deviation of this Gaussian to $W/2$. The final SVF $\bm{\Psi}$ can be integrated with the scale-and-square algorithm to obtain the final deformation field~\cite{arsigny2006log}; the inverse field can be obtained by integrating the negated SVF ($-\bm{\Psi}$).

\subsection{Implementation details}
The CNN is a standard U-net~\cite{ronneberger2015u}, with a design inspired by nnU-net~\cite{isensee2021nnu}. It has four resolution levels with two convolutional layers (comprising $3\times3\times3$ convolutions and a ReLu) followed by $2\times2\times2$ max pooling (in the encoder) or upconvolution (decoder). The final activation layer is linear, to regress the atlas coordinates in decimeters (which roughly normalizes them from -1 to 1). 
In addition to predicting atlas coordinates, the CNN also learns to predict a brain mask, which is used to define the domain $\Omega$  (i.e., to mask the loss during training, and to select the voxels that are used to fit the transforms at inference). For this, we add a segmentation loss to Equation~\ref{eq:loss} with relative weight 0.5; the segmentation loss itself combines a Dice loss (weight 0.75) and cross-entropy loss (weight 0.25). The model is trained with stochastic gradient descent with learning rate 0.01, weight decay $3\mathrm{e}{-5}$, and momentum 0.99 for 100 epochs, setting aside 20\% of the data for validation (selecting the best model). Input images are normalized with their median value inside the brain mask. At test time,  nonlinear models are fitted on the residual of the Affine-RANSAC fit (clipped at 10mm for outlier rejection). The RANSAC algorithm uses 50,000 randomly selected voxels, a maximum of 100 iterations, tests 500 voxels in every attempt, and requires 20\% of voxels to be inliers to consider any given solution. The minimum number of voxels per supervoxel in the log-polyaffine model is 100.


\section{Experiments and Results}

The training data consists of high-resolution, isotropic, T1-weighted scans of 897 subjects from the HCP dataset~\cite{van2013wu} and 1148 subjects from the ADNI~\cite{jack2008alzheimer}. 
The test data consists of high-resolution, isotropic, T1 of the first 100 subjects from both the ABIDE~\cite{di2014autism} and OASIS3~\cite{lamontagne2019oasis} datasets, for a total of 200 test subjects. All scans were segmented into 36 regions using FreeSurfer~\cite{fischl2002whole}. Details on the acquisitions can be found in the corresponding publications.

\subsection{Data preparation for RbR training}
The scans in the training dataset were masked and registered to the ICBM 2009b Nonlinear Symmetric MNI template using NiftyReg~\cite{modat2010fast}. Specifically, we first ran the block matching algorithm for affine registration (\texttt{reg\_aladin}) with the \texttt{-noSym} option, and subsequently used the fast free-form deformation algorithm (\texttt{f3d}) to compute the nonlinear registration. \texttt{f3d} was run in diffeomorphic mode (\texttt{-vel}) and local normalized correlation coefficient ($\sigma=5$) as similarity metric. The processing time for the whole dataset was less than 24h on a 64-core desktop.

\subsection{Experimental setup}

We quantify registration accuracy with average Dice scores between the segmentations of the scans and the segmentation of the atlas, deformed with the estimated registrations. We further evaluate the regularity of the deformations using the membrane energy, which is one of the smoothness metrics used in the registration literature. Using these metrics, we compared RbR with:

\smallskip\noindent\textbf{NiftyReg:}
We use the same parameters as for the training data.
Since we used it as ground truth in training, NiftyReg provides a ceiling for the performance that RbR can achieve. 

\smallskip\noindent\textbf{KeyMorph~\cite{wang2023robust}:}
Our main competing method is KeyMorph. We test KeyMorph variants as trained by the original authors (2024-03-06 version). For affine registration, we use 128 keypoints with weights optimized for affine Dice overlap. To estimate non-linear transforms, we use the TPS model with 512 keypoints, Dice-specific weights, and $\lambda_\mathrm{TPS} = 0$ (which yields the highest Dice in their article). We intensity normalize and skull-strip scans as in the original publication.

\smallskip\noindent\textbf{SynthMorph~\cite{hoffmann2021synthmorph}:} for completeness, we also consider a non-interpretable state-of-the-art algorithm trained to maximize segmentation overlap (Dice scores) with synthetic images. SynthMorph has an affine model based on keypoints~\cite{hoffmann2023affine} and a nonlinear model based on regression~\cite{hoffmann2023joint}. We finetuned SynthMorph with the MNI atlas permanently selected as moving image, and evaluated its affine and nonlinear modules separately. We used its default trade-off parameter $\lambda = 0.5$.

\begin{figure}[t!]
\centering
    \includegraphics[width=0.98\textwidth]{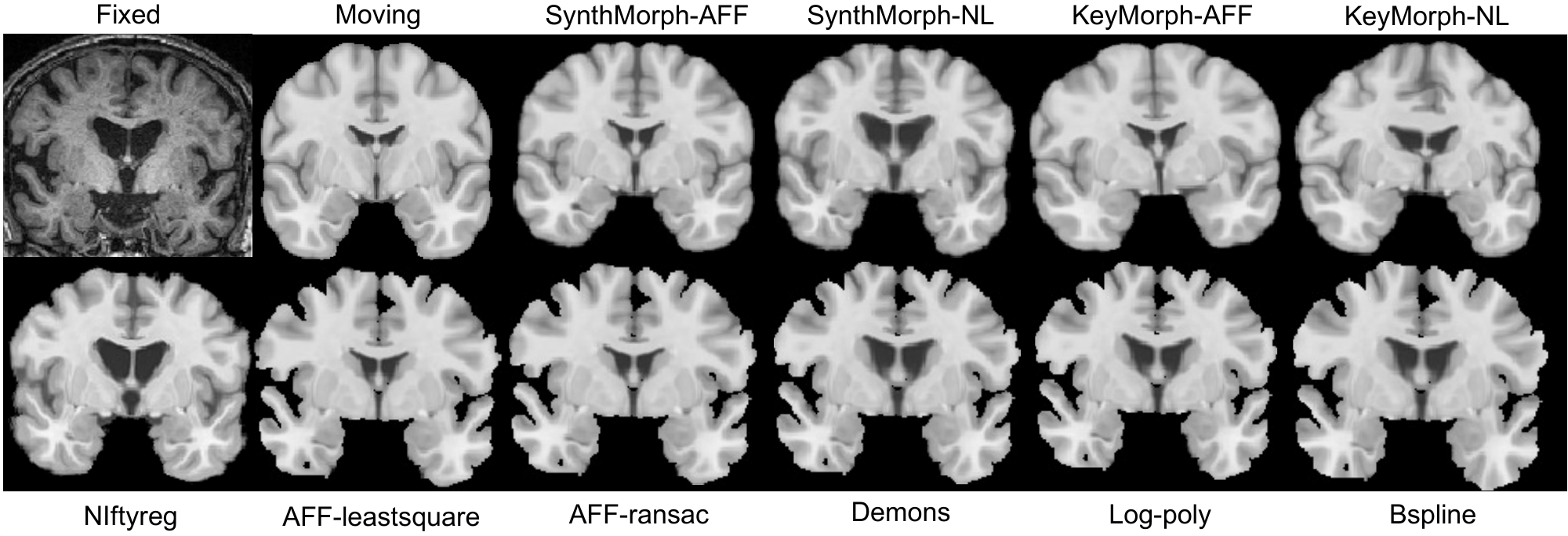}
  \caption{Coronal slice of sample fixed image and corresponding registered MNI slice, using the different approaches.}
  \label{fig:examples}
\end{figure}

\begin{figure}[t!]
\centering
    \includegraphics[width=0.98\textwidth]{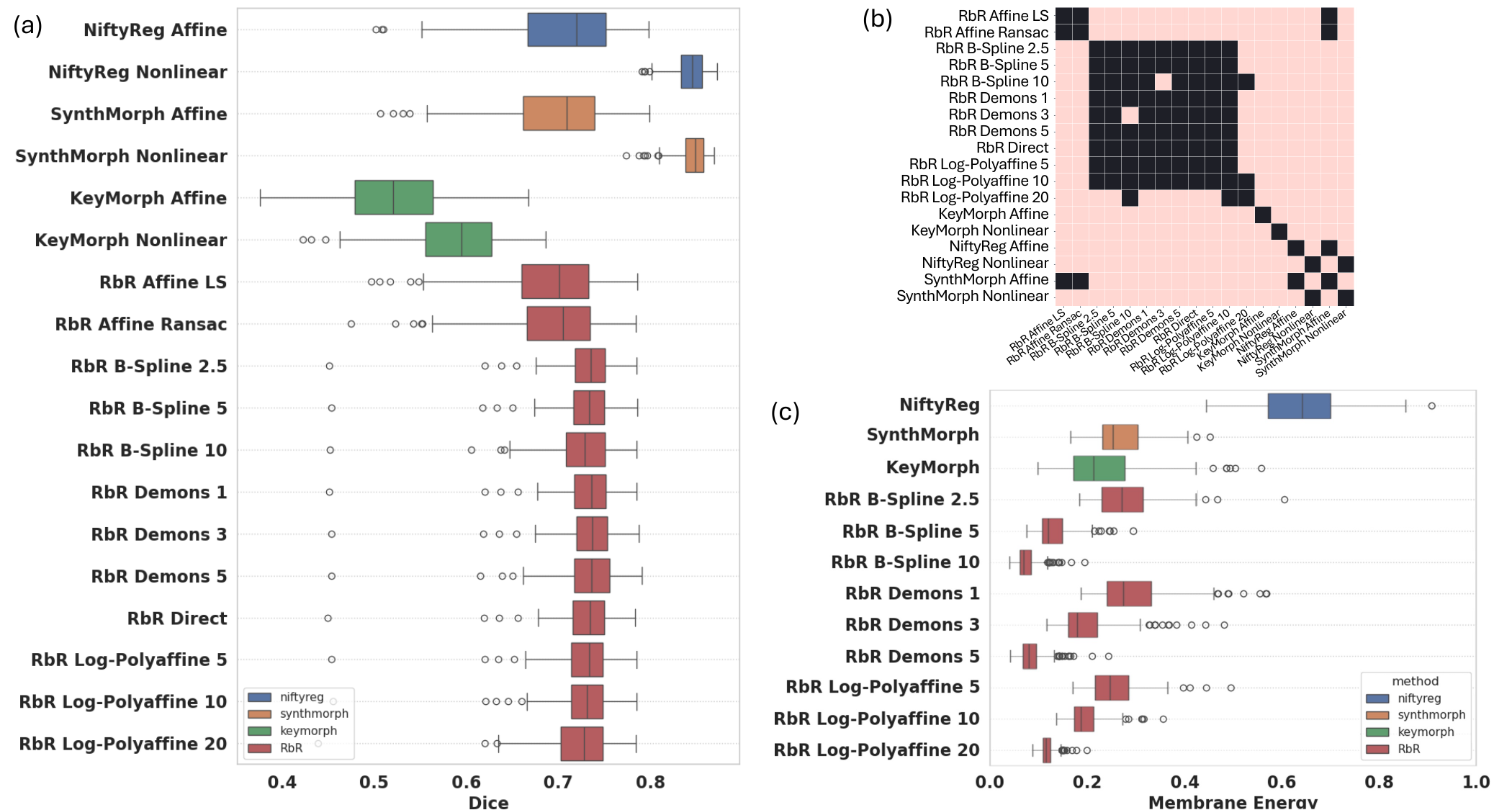}
  \caption{(a) Box plot of average Dice between segmentations of scans and registered MNI. For RbR, the numbers indicate control point spacing (Bsplines), Gaussian $\sigma$ (demons), or supervoxel width (log-polyaffine). (b)~Significance matrix for two-tailed t-tests comparing the Dice scores; pink represents p$<$0.05. (c)~Membrane energy for nonlinear models. In the box plots, the center line is the median, the ends of the box are the 25$^{th}$ and 75$^{th}$ percentiles, and the whiskers extend to the furthest observations not considered outliers -- which are marked with circles. Outliers are values more than 1.5 times the interquartile range away from the ends of the box. }
  \label{fig:dice}
\end{figure}

\subsection{Results}

Figure~\ref{fig:examples} shows qualitative registration results, while Figure~\ref{fig:dice} shows box and significance plots for the average Dice and membrane energy. KeyMorph with TPS yields an average Dice of 0.60, which is 7 points lower than the value reported in the original publication. This could be a domain gap issue -- including both the test datasets and the MNI atlas, which is blurrier than a regular scan. Errors are apparent in Figure~\ref{fig:examples}, where KeyMorph fails to correct the rotation about the A-P axis in linear mode and introduces wrong deformations in nonlinear mode.

Compared with KeyMorph, RbR does not require skull-stripping and yields better Dice, with medians of $\sim$0.70 (affine) and $\sim$0.73 (nonlinear) --  higher than Keymorph's $\sim$0.60 on our dataset or even the value reported in their publication ($\sim$0.67). While the improvement from linear to nonlinear models in RbR may not be large in terms of median, it is very noticeable in terms of the first quartile and outliers. In Figure~\ref{fig:examples}, the difference between the nonlinear versions is subtle, but noticeable e.g., around the third ventricle (less aggressively ``closed'' by the Bspline model). The ability to more accurately follow contours is paid in terms of membrane energy, which is noticeably higher for the more flexible models.

Finally, we note that, while RbR-affine and Synthmorph-affine (both based on keypoints) achieve the same Dice as NiftyReg in the affine model, keypoint methods still trail non-interpretable approaches (classical and learning-based) in nonlinear registration: NiftyReg and SynthMorph achieve Dice scores 10 points higher than RbR and $\sim$20 points higher than KeyMorph in this setting -- and SynthMorph does so with similar levels of membrane energy.


\subsection{Discussion and conclusion}
RbR offers a new perspective on interpretable keypoint-based registration, posing it as a coordinate regression problem. Allowing every voxel to inform the model fit enables RbR to outperform standard keypoint approaches and fit less parsimonious nonlinear models. RbR also has disadvantages, particularly the need to retrain the model for each new atlas. Future work will include fitting other models, adding topological losses in training (e.g., penalizing negative Jacobians), investigating RbR's value as a feature extractor (i.e., as pretraining for other tasks),  exploring other improvements that help close the gap with non-interpretable approaches, and integrating with domain randomization approaches for contrast- and resolution-agnostic registration (as in~\cite{hoffmann2023affine}). We believe that RbR will be particularly valuable in scenarios requiring robust and interpretable registration in presence of extreme rotations, such as fetal imaging.



%
%

\bibliographystyle{splncs04}
\bibliography{Reference}

\end{document}